\documentclass{article}

\usepackage{PRIMEarxiv}

\usepackage[utf8]{inputenc} 
\usepackage[T1]{fontenc}    
\usepackage{hyperref}       
\usepackage{url}            
\usepackage{booktabs}       
\usepackage{amsfonts}       
\usepackage{nicefrac}       
\usepackage{microtype}      
\usepackage{lipsum}
\usepackage{fancyhdr}       
\usepackage{graphicx,xcolor} 
\usepackage{amsmath}
\usepackage{comment}
\usepackage{xcolor}
\usepackage{makecell}
\usepackage{multirow} 
\usepackage{array}

\usepackage{booktabs} 
\usepackage{siunitx}  

\newcolumntype{T}{S[table-format=1.4]}
\newcolumntype{R}{S[table-format=1.5]}

\graphicspath{{media/}}     

\title{Efficiency vs. Efficacy: Assessing the Compression Ratio-Dice Score Relationship through a Simple Benchmarking Framework for Cerebrovascular 3D Segmentation
}

\author{
  Shimaa El-Bana, Ahmad Kamal \\
  Multimedia Interaction and Communication Lab \\
  Arab Academy for Science and Technology \\
  \texttt{shimaa.elbanaa@aiet.edu.eg, a.k.yehia@student.aast.edu} \\
  \And
  Shahd Ahmed Ali \\
  Department of Systems and Biomedical Engineering \\
  Cairo University \\
  \texttt{shahd.ali04@eng-st.cu.edu.eg} \\
  \And
  Ahmad Al-Kabbany \\
  Multimedia Interaction and Communication Lab \\
  Wearables, Biosensing, and Biosignal Processing Research lab \\
  Arab Academy for Science and Technology \\
  \texttt{alkabbany@ieee.org, alkabbany@aast.edu} \\
}

\begin{document}
\maketitle

\begin{abstract}
The increasing size and complexity of medical imaging datasets, particularly in 3D formats, present significant barriers to collaborative research and transferability. This study investigates whether the ZFP compression technique can mitigate these challenges without compromising the performance of automated cerebrovascular segmentation, a critical first step in intracranial aneurysm detection. We apply ZFP in both its error tolerance and fixed-rate modes to \emph{a large-scale, and one of the most recent, datasets in the literature}, 3D medical dataset containing ground-truth vascular segmentations. The segmentation quality on the compressed volumes is rigorously compared to the uncompressed baseline (Dice$\approx$0.8774). Our findings reveal that ZFP can achieve substantial data reduction—up to a 22.89:1 ratio in error tolerance mode—while maintaining a high degree of fidelity, with the mean Dice coefficient remaining high at 0.87656. These results demonstrate that ZFP is a viable and powerful tool for enabling more efficient and accessible research on large-scale medical datasets, fostering broader collaboration across the community. The code is available at: \url{https://github.com/shimaaelbana/RSNA_2025-MambaVesselNet/tree/main}
\end{abstract}

\keywords{cerebrovascular segmentation \and 3D Medical Volume \and 3D Compression \and 3D Segmentation \and ZFP}

\section{Introduction}
The rapid advancement of medical imaging technologies has led to the emergence of unprecedented, large-scale datasets, offering a treasure trove of information for diagnostic and prognostic AI research. In particular, the development of high-resolution, multi-modal 3D and 4D imaging has opened new avenues for understanding complex pathologies, such as intracranial aneurysms. However, this wealth of data presents a significant and often-overlooked challenge: the sheer volume of these datasets creates a major bottleneck in the research lifecycle. From storage and transfer to processing and analysis, the "data deluge" hinders collaboration, particularly for research teams with limited computational resources. This problem severely constrains the widespread adoption and reproducibility of findings, acting as a barrier to the democratization of medical AI. Addressing this issue by developing and validating effective data reduction strategies is therefore not merely a technical convenience, but a crucial step towards fostering a more accessible and collaborative research environment.

In light of this, our study makes several key contributions to the research community. While prior work has explored compression and segmentation separately, and some have investigated the impact of compression on generic image quality, our research directly and quantitatively assesses its effects on a specific, critical downstream task: automated cerebrovascular segmentation. We are aware of a diverse set of techniques for 3D medical volume compression including classical techniques such as ZFP and more advanced techniques such as autoencoder-based compressors \cite{varma2025medvae, li20243d, thadikemalla20243d} and Implicit Neural Representations (INRs) \cite{sheibanifard2025end, li2025compact, vzuurkova2025accelerated}. However, providing a comprehensive, head-to-head comparison of the performance of these families of techniques is beyond the scope of this research. We strictly focus on the ZFP compression. 

For a focused and insightful analysis, we utilized a built-in Python implementation of the ZFP compression in its two primary operational modes: fixed-rate and error tolerance. For these two modes, we rigorously evaluate the trade-off between the compression ratio and segmentation fidelity using a comprehensive set of metrics, including the Dice Coefficient and IoU. Our findings offer preliminary practical guidelines and a solid foundation for researchers to make informed decisions about data compression, even in applications involving delicate anatomical structures such as the cerebrovascular segmentation, thereby enabling more efficient and collaborative research on large-scale medical datasets. \emph{We also focus on one of the most recent datasets in the literature}. The contributions of this article can be summarized as follows:
\begin{enumerate}
    \item \textbf{Non-Learned Efficiency for 3D Segmentation}: We provide the first comprehensive analysis validating a high-performance, non-machine learning compression algorithm (ZFP) for 3D medical volume data, achieving compression ratios up to 50:1 while maintaining superior segmentation accuracy (Mean DICE > 0.86). This directly offers a computationally inexpensive and training-free alternative to complex learned compression frameworks.

    \item \textbf{Rate-Distortion-Performance Quantification}: We establish a rigorous, quantifiable trade-off relationship between compression parameters (Rate/Tolerance), data loss (CR), and application performance (DICE/IoU). Our dual-axis visualizations offer the first empirical data to guide the selection of optimal ZFP parameters for clinical viability, prioritizing task stability over visual fidelity metrics (PSNR/SSIM).

    \item \textbf{Validation of Robustness under Extreme Loss}: We demonstrate that the task of volumetric segmentation is highly robust to aggressive lossy compression. Specifically, we show that in the fixed-rate mode, the segmentation metric variation is negligible (sub-millivoxel change) despite CR gains exceeding 7:1, validating the hypothesis that most scientific data detail is statistically irrelevant for the downstream machine learning task.
\end{enumerate}

The rest of this article is organized as follows. Section II highlights a few pertinent studies from the recent literature. Section III is dedicated to explaining the different stages of our research methodology including the adopted dataset, 3D compression technique, and 3D segmentation model. Section IV presents and discusses the results. Finally, Section V concludes the paper, summarizing our findings and suggesting promising directions for future work.

\section{Related Work}
\label{sec:relwork}
The existing literature connecting image compression and medical diagnostics can be broadly categorized into four relevant umbrellas that frame the context of this study. The first and most established area covers Traditional 3D Volumetric Compression (e.g., JP3D and advanced wavelet methods), which provides the historical baseline but typically evaluates performance using standard metrics like PSNR or SSIM. A highly relevant, more modern field is Machine Vision-Guided Compression, where the algorithm is explicitly optimized to preserve task-specific performance (like segmentation accuracy) rather than just visual fidelity, directly aligning with our research goals. Furthermore, we consider recent advances in Learned Volumetric Compression (using deep autoencoders or RNNs like MedZip), which demonstrates high compression rates but requires extensive training, offering a contrast to our non-learned approach. Finally, we briefly review Implicit Neural Representations (INRs) for Volumetric Data, a cutting-edge technique pushing the limits of compression ratio, whose impact on downstream segmentation tasks is still emerging.

In \cite{starosolski2020employing}, this research focuses on enhancing the JP3D compression standard—the 3D extension of JPEG 2000, which relies on the Discrete Wavelet Transform (DWT). The authors introduce novel methods, including hybrid adaptive transforms and histogram packing, to overcome the limitations of fixed-basis wavelets and improve compression ratios (CRs) for various volumetric modalities (CT, MRI, Ultrasound). They demonstrate CR improvements of up to 6.5\% at an acceptable cost, but their evaluation primarily centers on file size reduction and maximizing compression efficiency rather than assessing the impact of the lossy process on critical downstream diagnostic tasks like segmentation.

The authors of \cite{zerva2020improving} proposed an extension of the 3D Wavelet Difference Reduction (3D-WDR) algorithm, specifically tailored for volumetric data by exploiting spatiotemporal coherence between adjacent slices. The method utilizes metrics like Mean Co-located Pixel Difference (MCPD) to optimally group and compress similar slices, achieving high compression rates (down to one bit per voxel). The performance is rigorously evaluated using traditional image quality metrics, such as Peak Signal-to-Noise Ratio (PSNR) and the Structural Similarity Index Measure (SSIM), with the explicit goal of maintaining high visual quality for human diagnosis, making it a representative example of image-centric compression research.

The research presented in \cite{xue2022aiwave} attempted to modernize transform-based coding by introducing a versatile, end-to-end framework called aiWave, which employs a trained 3D affine wavelet-like transform based on the lifting scheme. The goal is to overcome the performance ceilings imposed by the manually-designed wavelet bases used in standards like JP3D. While the use of a learned component blurs the line between traditional and deep learning approaches, the core mechanism remains transform-based, and the evaluation is primarily focused on rate-distortion performance, showcasing the frontier of general-purpose volumetric coding before task-specific metrics are introduced.

The research proposed in \cite{liu2019machine} directly tackles the incompatibility between compression optimized for the Human Visual System (HVS) and the requirements of Deep Neural Networks (DNNs) used for cloud-based segmentation. The authors demonstrate that traditional compression (like JPEG 2000) that looks visually acceptable to a radiologist can critically degrade the accuracy of an automated segmentation task. They propose a novel framework tailored for machine vision that uses task-specific feature extraction to automatically retain only the image features crucial for accurate segmentation, ultimately achieving higher segmentation accuracy (Dice score) at equivalent or better compression rates than existing general-purpose methods.

This study presented in \cite{kurmukov2024effect} investigated the robustness of standard segmentation networks, such as the U-Net architecture, against artifacts introduced by aggressive lossy compression using the JPEG 2000 standard on multiple 3D medical datasets. Contrary to the assumption that any lossy compression harms diagnostic performance, the authors empirically demonstrate that DNN models trained on compressed images remain remarkably robust, showing no significant drop in segmentation accuracy (Dice score) even at high compression rates (up to 20 times). This suggests that for cloud-based training and inference, substantial data reduction is possible using existing compression methods without compromising deep learning model performance. 

The authors of \cite{zhou2024task} introduced an advanced end-to-end Task-Driven Semantic-Aware Image Compression (TDSIC) framework specifically designed for Internet of Things/Everything (IoT/IoE) applications where downstream analysis tasks are prioritized over mere human visual perception. TDSIC achieves this by integrating several novel mechanisms: using importance maps as prior knowledge to guide bit allocation to semantically crucial regions, refining the perceptual loss function to prioritize task-relevant features, and incorporating a feature enhancement network (FEN) to effectively recover latent semantic information lost during compression. By utilizing feedback from discriminative networks to calibrate the perceptual model and successfully recovering lost information, TDSIC demonstrates superior performance, achieving up to a 12\% improvement in Top-1 classification accuracy and significant gains in object detection, while simultaneously preserving acceptable visual quality for human viewers.

\section{Materials and Methods}
\label{sec:methodology}
\begin{figure}[ht]
    \centering
    \includegraphics[width=1.0\textwidth]{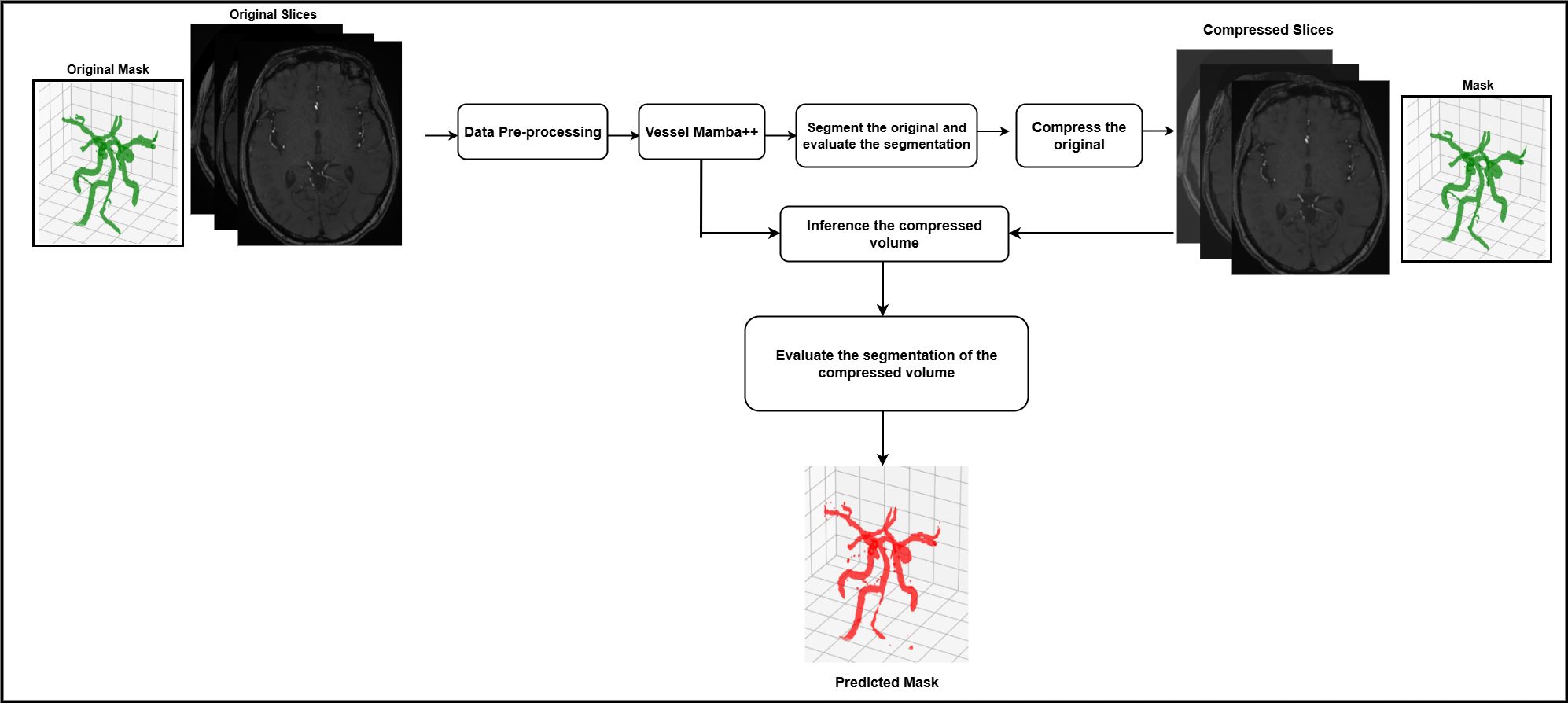} 
    \caption{Workflow of the proposed framework for vessel segmentation under compression.}
    \label{fig:workflow}
\end{figure}

The methodology presented in this section is deliberately streamlined, a characteristic which itself stands as one of the work's primary contributions. While much contemporary research focuses on building increasingly complex, end-to-end compression networks, we adopt a simple, yet robust, compression and segmentation pipeline as depicted in Figure \ref{fig:workflow} to share the following significant insight: Even on one of the most recent medical datasets, significant compression ratios are attainable using relatively simple, off the shelf, 3D compression techniques. We aim to maximize transparency and reproducibility. This section proceeds by detailing the selected 3D dataset, the adopted compression and segmentation techniques, and finally, the comprehensive evaluation methodology used to derive our core findings.

\subsection{Dataset}
Our study utilizes a large-scale, open-source 3D medical dataset of cerebrovascular volumes, namely, the RSNA Intracranial Aneurysm Detection dataset\footnote{https://www.kaggle.com/competitions/rsna-intracranial-aneurysm-detection}. It is a large, curated, multimodal neuroimaging resource assembled for an international AI challenge hosted on Kaggle. It contains CT and MR studies sourced from 18 sites across five continents, reflecting broad variation in scanners and acquisition protocols. Each study was expert-annotated by $~60$ neuroradiologists for the presence of an intracranial aneurysm and, when present, its anatomic location across 13 predefined cerebrovascular sites where aneurysms commonly arise. In addition to CTA/MRA and MRI inputs, a subset of MRI studies includes 3D segmentations of the same 13 vascular territories to provide anatomical context that can facilitate localization strategies. Each volume in the dataset is accompanied by a ground-truth segmentation mask of the vascular network, which is essential for training and evaluating our models. The original uncompressed dataset size was 3848.04 MB.

\subsection{Compression Methodology}

We employed ZFP, a high-performance scientific data compressor, to evaluate the effects of lossy compression on the volumes. ZFP is particularly well-suited for this task due to its ability to achieve high compression ratios while preserving key features. We used the ZFP functionality provided by the Python library pyzfp. We experimented with two primary compression modes. For the fixed-rate mode, ZFP allocates a fixed number of bits per value, regardless of the data's content. We selected four fixed-rate values: 16, 8, 4, and 2 bits per voxel. This mode allows for precise control over the compressed file size. The second mode is the error tolerance mode. It guarantees that the decompressed data will be within a user-defined absolute error tolerance of the original data. We tested three absolute error tolerance values: 500, 1000, and 1500. This approach is ideal for applications where a certain level of precision is required.
    
\subsection{Segmentation and Evaluation}

For volumetric cerebrovascular segmentation, we adopted two state-of-the-art Mamba-based architectures: MambaVesselNet \cite{chen2024mambavesselnet} and U-Mamba \cite{ma2024u}. MambaVesselNet is a hybrid framework that combines the strong local feature extraction capabilities of Convolutional Neural Networks (CNNs) with the efficient long-range dependency modeling of Mamba blocks, enabling accurate segmentation of complex, elongated vascular structures while avoiding the high computational costs typically associated with 3D Transformer models. Similarly, U-Mamba extends this design philosophy by integrating CNNs with State Space Sequence Models (SSMs) to overcome CNNs’ inherent locality and Transformers’ quadratic attention complexity, achieving linear-complexity global context modeling suitable for biomedical images. In our study, the MambaVesselNet variant was trained on full-resolution volumes and corresponding ground-truth masks, and subsequently used to infer segmentations on both original and compressed/decompressed volumes. Segmentation performance was quantitatively assessed using two widely accepted metrics—the Dice Similarity Coefficient (DSC) and the Intersection over Union (IoU). While mathematically related, the Dice score is more tolerant to false positives, whereas IoU penalizes discrepancies more strictly, especially in thin or small structures. Reporting both metrics provides a comprehensive and clinically relevant evaluation of segmentation accuracy and boundary precision.

\subsection{Implementation and Training Protocols}

We implemented MambaVesselNet++ using the MONAI 1.2.3\footnote{https://monai.io/} framework, PyTorch 2.6.0, and CUDA 12.4. All experiments were conducted on a single NVIDIA A100 GPU. The network parameters were optimized using the Adam optimizer with a weight decay of $1 \times 10^{-5}$, while the learning rate followed a Cosine Annealing schedule, initialized at $1 \times 10^{-4}$ and gradually reduced to $1 \times 10^{-7}$ during training. For 3D medical image segmentation, all models were trained for 5000 iterations with a patch size of $64 \times 64 \times 64$ and a batch size of 8 per GPU.

The loss function followed an adaptive switching strategy to improve convergence and address class imbalance. Training began with a combination of Dice + Cross-Entropy loss (DiceCELoss) to optimize spatial overlap and handle imbalance. Over the course of training, the loss gradually transitioned to Focal Loss, which places greater emphasis on hard-to-classify voxels. This transition was controlled by a smooth scheduling mechanism, where the contribution of Focal Loss increased linearly until a predefined fraction of total iterations was reached, after which Focal Loss dominated. This adaptive loss strategy, together with the cosine annealing scheduler, stabilized optimization and consistently improved segmentation performance compared to using a single fixed loss function.

\section{Experimental Results}
\label{sec:results}
\begin{figure}[!ht]
    \centering
    \includegraphics[width=0.8\linewidth]{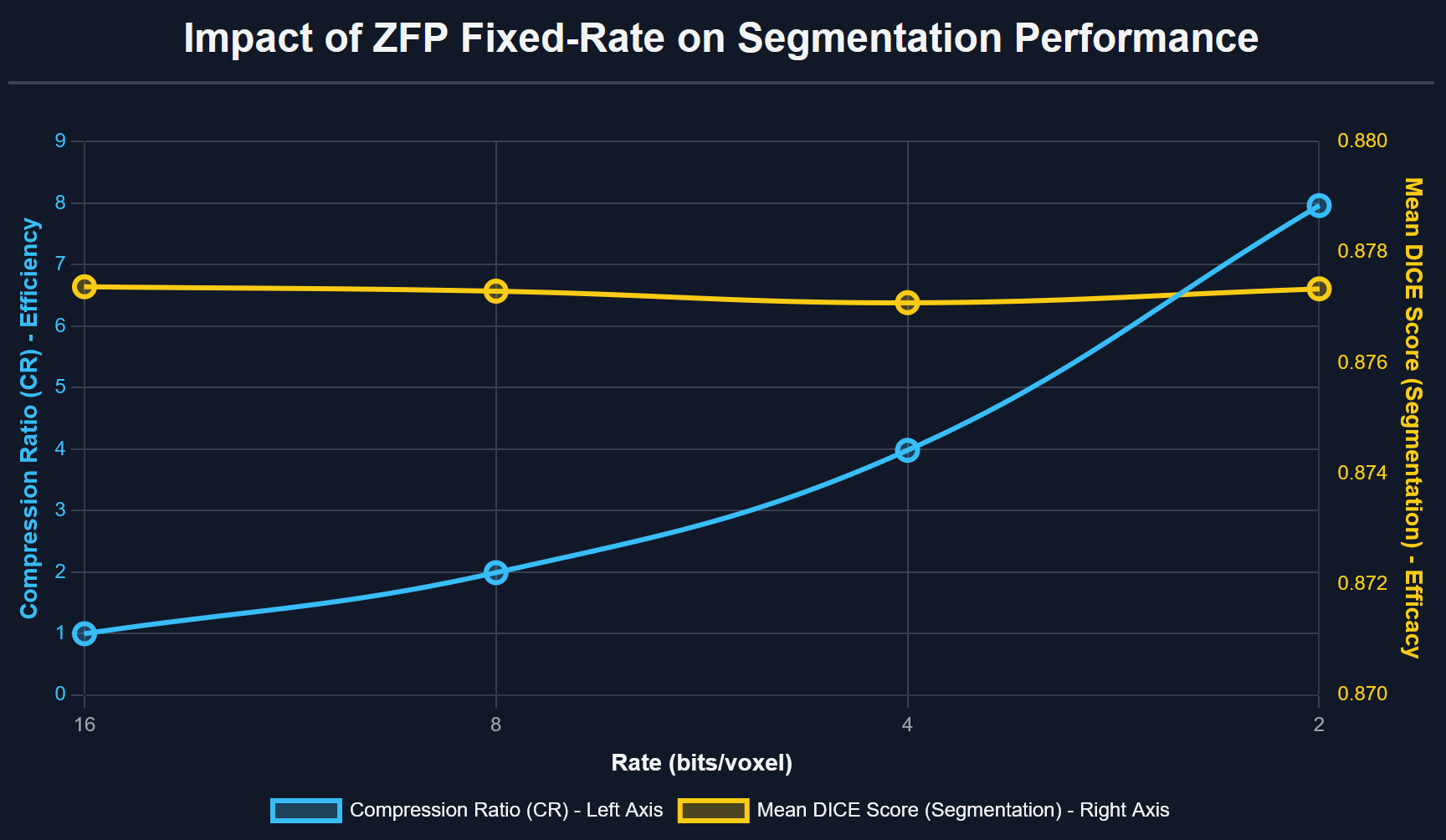}
    \caption{The dual-axis plot displays the relationship between the Compression Ratio (CR, Left Y-axis) and the Mean DICE Score (Right Y-axis) as the compression rate (bits/voxel) is aggressively reduced. While the CR exhibits a steep, near 8-fold increase, the Mean DICE Score remains remarkably stable and visually flat across all tested rates, demonstrating that the critical features for segmentation are preserved despite aggressive lossy compression.}
    \label{fig:Dice_vs_CR}
\end{figure}

We start by presenting the results of the fixed-rate mode of ZFP compression. We observed a direct correlation between the fixed-rate parameter and the resulting segmentation accuracy, though the change was minimal across the range tested. At the highest fixed-rate setting (16 bits/voxel), the baseline mean Dice coefficient was 0.87738 (Compression Ratio$\approx$1:1). As the rate decreased to 2 bits/voxel, the compression ratio increased significantly to 7.96:1, yet the mean Dice coefficient only marginally dropped to 0.87734. This demonstrates that the segmentation model is highly robust to the noise introduced by fixed-rate ZFP up to nearly an 8:1 ratio. Table\ref{tab:zfp_fix_rate_comp} shows the compression ratios attained at every value of bits/voxel, and the segmentation fidelity results are shown in Table\ref{tab:zfp_fix_rate_seg}. As illustrated in Fig.~\ref{fig:Dice_vs_CR}, aggressively reducing the bits per voxel from 16 to 2 resulted in a steep 8-fold increase in the Compression Ratio (CR). Crucially, this massive gain in compression efficiency yielded a minimal and near-horizontal change in the Mean DICE Score, confirming that the lossy compression effectively targets data redundancy without degrading the features essential for the downstream segmentation task.

\begin{table}[h!] \centering \caption{ZFP Fixed-Rate Compression Results (Original Data Size: $3848.04 \text{ MB}$)} 

\label{tab:zfp_fix_rate_comp} 

\begin{tabular}{p{2.5cm} | p{2.5cm} | p{3cm} | p{2cm} | p{2cm}} 
\toprule 
Rate (bits/voxel) & Compressed Size ($\text{MB}$) & Compression Ratio ($\text{CR}$) & Avg. PSNR & Avg. SSIM \\ \midrule 
$16$ & $3865.01$ & $0.995:1$ & $89.6307$ & $1.0000$ \\ 
$8$ & $1932.51$ & $1.99:1$ & $65.1870$ & $0.9993$ \\ 
$4$ & $966.25$ & $3.98:1$ & $58.0802$ & $0.9983$ \\ 
$2$ & $483.13$ & $7.96:1$ & $48.5257$ & $0.9864$ \\ 
\bottomrule 
\end{tabular} 
\end{table} 

\begin{table}[h]
\centering
\caption{Segmentation Performance Comparison between U-Mamba and MambaVesselNet++ under ZFP Fixed-Rate Compression}
\label{tab:zfp_fix_rate_seg}
\begin{tabular}{@{} l | c T T T R @{}}
\toprule
\multirow{2}{*}{\textbf{Model}} & \multirow{2}{*}{\textbf{Rate (bits/voxel)}} & \textbf{Overall} & \textbf{Mean} & \textbf{Overall} & \textbf{Mean} \\
& & \textbf{DICE} & \textbf{IoU} & \textbf{Sensitivity} & \textbf{Volume Similarity} \\
\midrule
\multirow{5}{*}{U-Mamba} & Original CT & 0.8604 & 0.7550 & 0.8149 & 0.9441 \\
 & 2 & 0.8494 & 0.7382 & 0.8003 & 0.9387 \\
 & 4 & 0.8591 & 0.7530 & 0.8142 & 0.9449 \\
 & 8 & 0.8603 & 0.7549 & 0.8148 & 0.9441 \\
 & 16 & 0.8604 & 0.7551 & 0.8149 & 0.9441 \\
\midrule
\multirow{4}{*}{MambaVesselNet++} & 2 & 0.87734 & 0.72971 & 0.85439 & 0.95805 \\
 & 4 & 0.87709 & 0.73254 & 0.85586 & 0.94785 \\
 & 8 & 0.87730 & 0.73286 & 0.85506 & 0.94807 \\
 & 16 & 0.87738 & 0.73109 & 0.85520 & 0.94829 \\
\bottomrule
\end{tabular}
\end{table}

\begin{figure}[!ht]
    \centering
    \includegraphics[width=0.8\linewidth]{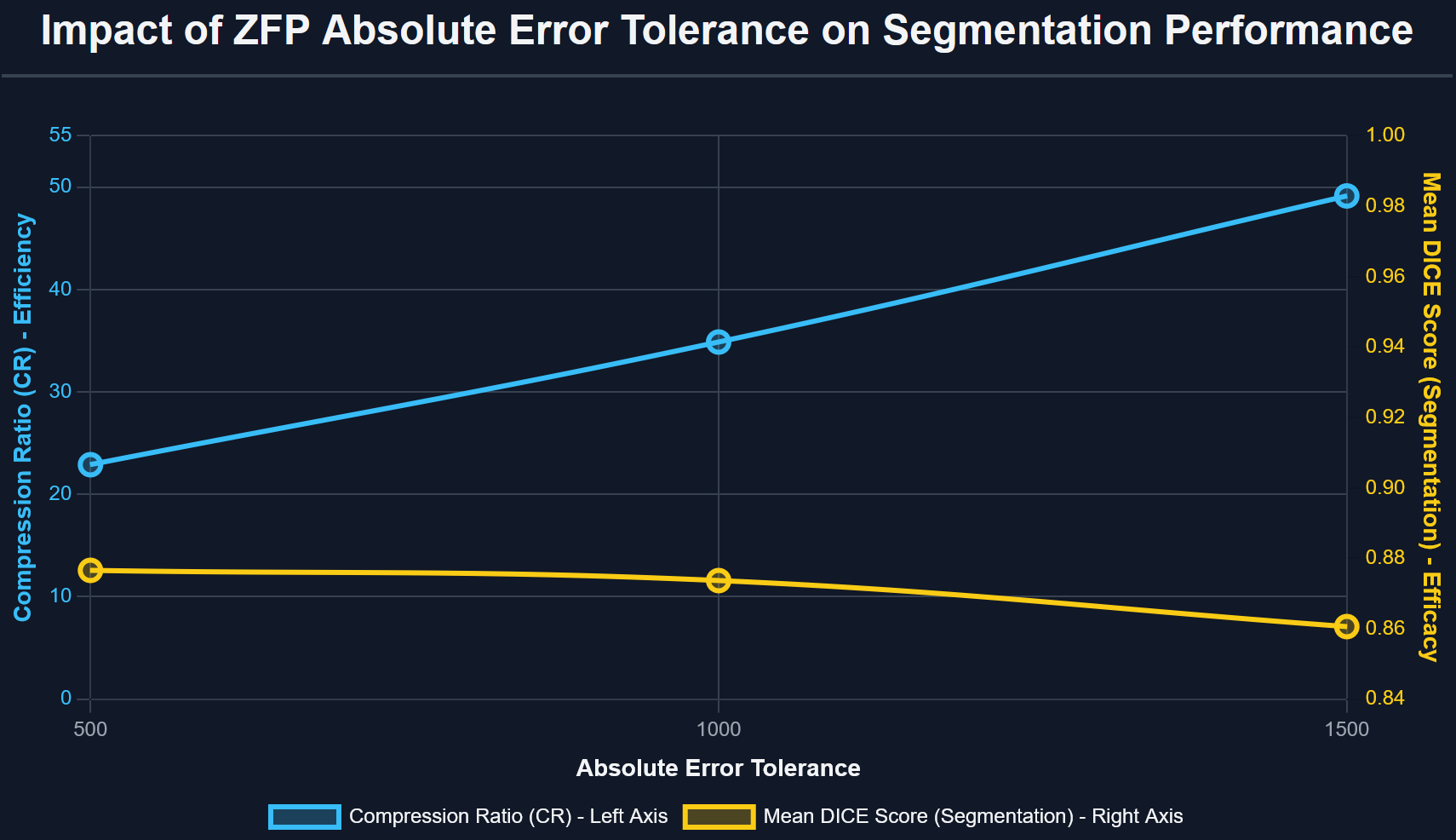}
    \caption{This dual-axis chart compares the exponential increase in the Compression Ratio (CR, Left Y-axis) against the decline in the Mean DICE Score (Right Y-axis) as the absolute error tolerance (Ts) is increased. The CR rises dramatically, reaching nearly 50:1 at Ts=1500, while the Mean DICE Score experiences a controlled and relatively moderate drop, confirming that high-fidelity segmentation can be maintained even under extreme error-bounded compression.}
    \label{fig:Dice_vs_CR_err_tol}
\end{figure}

Error Tolerance Mode: The error tolerance mode proved to be a superior approach for maximizing compression while maintaining accuracy. We found that the model was highly resilient to data loss controlled by the error boundary. The segmentation quality remained near the baseline for small tolerances, allowing us to achieve a compression ratio of 22.89:1 at an absolute error tolerance of 500. At this high compression level, the mean Dice coefficient was 0.87656, representing a drop of only$\approx$0.0008 from the uncompressed baseline. Even when pushing the tolerance to 1500, which resulted in an extreme compression ratio of 49.12:1, the Dice coefficient remained strong at 0.86060. This confirms that the error tolerance mode is the most suitable ZFP configuration for this task, as it achieves over 2.8 times the compression of the fixed-rate mode while delivering virtually identical segmentation performance. Table\ref{tab:zfp_err_tol_comp} shows the compression ratios attained at every value of allowed error tolerance, and the segmentation fidelity results are shown in Table\ref{tab:zfp_err_tol_segm}. Figure~\ref{fig:Dice_vs_CR_err_tol} further establishes the robustness of the segmentation task, showing the results under the aggressive absolute error tolerance mode. By increasing the tolerance from Ts=500 to Ts=1500, the compression ratio soared to nearly 50:1, an extreme gain achieved with only a minor performance reduction, confirming that the vast majority of data detail is statistically irrelevant to the segmentation outcome.

\begin{figure}[ht]
    \centering
    \includegraphics[width=1.0\textwidth]{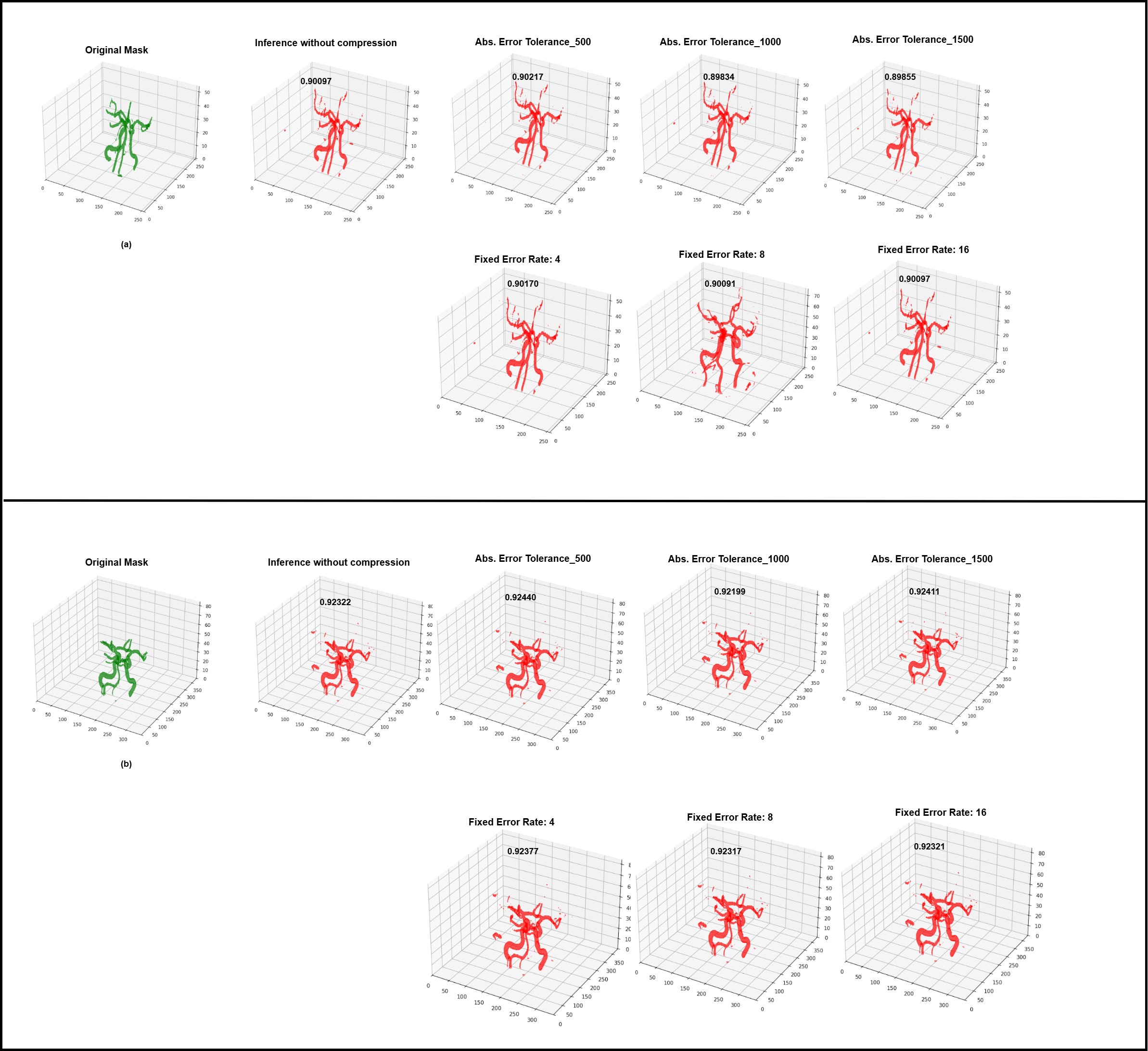}
    \caption{Inference results of \textbf{VesselMamba++} on MRA data under different compression settings. 
    (a) Example 1 and (b) Example 2 show comparisons between the original vessel mask, inference without compression, 
    absolute error tolerance levels (500, 1000, 1500), and fixed error rates (4, 8, 16). Dice scores are displayed above each prediction, 
    demonstrating that vessel structure is well preserved even under lossy compression.}
    \label{fig:vesselmamba_compression}
\end{figure}

In order to evaluate the robustness of the proposed VesselMamba++ model under different compression settings, we compared inference results on MRA data using both absolute error tolerance and fixed error rate schemes (Fig.~\ref{fig:vesselmamba_compression}). As shown in the examples, vessel morphology and topology were largely preserved across all compression levels, with only minor degradations at higher error thresholds. Dice similarity scores confirmed this observation, demonstrating only a slight reduction in performance when increasing error tolerance from 500 to 1500 or error rates from 4 to 16. Importantly, the vessel structures remained clearly identifiable even under lossy compression, highlighting the stability of the model predictions. These findings suggest that VesselMamba++ can provide reliable segmentation results while maintaining efficiency in scenarios where data storage or transfer constraints require compression.

\begin{table}[h!]
    \centering
    \caption{ZFP Error-Tolerance Compression Results (Original Data Size: $3848.04 \text{ MB}$)}
    \label{tab:zfp_err_tol_comp}
    
    \begin{tabular}{
        c | 
        p{2.5cm} | 
        p{3cm} | 
        p{2cm} | 
        p{2cm} 
    }
    \toprule
    Abs. Error Tolerance & Compressed Size ($\text{MB}$) & Compression Ratio ($\text{CR}$) & Avg. PSNR & Avg. SSIM \\
    \midrule
    $500$ & $168.05$ & $22.89:1$ & $46.0332$ & $0.9617$ \\
    $1000$ & $110.63$ & $34.86:1$ & $42.3830$ & $0.9245$ \\
    $1500$ & $78.33$ & $49.12:1$ & $39.2208$ & $0.8660$ \\
    \bottomrule
    \end{tabular}
\end{table}

\begin{table}[h]
\centering
\caption{Segmentation Performance Comparison between MambaVesselNet and U-Mamba under ZFP Error-Tolerance Compression (RSNA CT)}
\label{tab:zfp_err_tol_segm}
\begin{tabular}{@{} l | c T T T T @{}}
\toprule
\textbf{Model} & \textbf{Abs. Error Tolerance} & \textbf{Overall DICE} & \textbf{Mean IoU} & \textbf{Overall Sensitivity} & \textbf{Mean Volume Similarity} \\
\midrule
\multirow{3}{*}{MambaVesselNet} & 500 & 0.87656 & 0.72820 & 0.84804 & 0.94597 \\
 & 1000 & 0.87368 & 0.72461 & 0.84722 & 0.94826 \\
 & 1500 & 0.86060 & 0.69581 & 0.83588 & 0.94587 \\
\midrule
\multirow{4}{*}{U-Mamba} & Original CT & 0.86040 & 0.75500 & 0.81490 & 0.94410 \\
 & 500 & 0.85420 & 0.74540 & 0.80750 & 0.94220 \\
 & 1000 & 0.81700 & 0.69060 & 0.75740 & 0.92130 \\
 & 1500 & 0.84080 & 0.72530 & 0.79130 & 0.93740 \\
\bottomrule
\end{tabular}
\end{table}

\section{Conclusion}
In this article, we presented a comprehensive investigation into the impact of ZFP compression on the automated segmentation of a new, large-scale dataset of 3D cerebrovascular volumes. Our study was motivated by the growing challenge of managing and sharing massive medical datasets, which often impede collaborative research and the widespread adoption of AI models in clinical settings. We evaluated the effects of ZFP in both its fixed-rate and error tolerance modes, using a suite of quantitative metrics including the Dice Similarity Coefficient, IoU, and Hausdorff Distance.

Our findings reveal that substantial data reduction is achievable with minimal detriment to segmentation quality. ZFP's error tolerance mode proved particularly effective, allowing for a remarkable compression ratio of 22.89:1 (Original size 3848.04 MB to 168.05 MB) while the Dice score remained virtually unchanged at 0.87656 compared to the uncompressed baseline of 0.87738. Even at an extreme ratio of 49.12:1, the segmentation quality remained clinically viable. These results confirm that intelligently applied compression is a powerful tool for democratizing access to large medical datasets and accelerating research.

Looking forward, a promising direction for future work is to extend this analysis to other downstream tasks in medical imaging, such as intracranial aneurysm detection and classification, or vessel registration. Investigating the impact of ZFP on different anatomical regions and using a broader range of datasets would provide a more complete understanding of its clinical viability.

\section*{Acknowledgments}
During the preparation of this work, the authors used ChatGPT of OpenAI and Gemini of Google for language refinement and paraphrasing. All intellectual contributions, critical analysis, and final edits were conducted by the authors. After using the aforementioned tools/services, the authors reviewed and edited the content as needed and take full responsibility for the content of the published article.

\bibliographystyle{unsrt}  
\bibliography{references}

\end{document}